\def\year{2017}\relax
\documentclass[letterpaper]{article} 
\usepackage{aaai17}  
\usepackage{times}  
\usepackage{helvet}  
\usepackage{courier}  
\usepackage{url}  
\usepackage{graphicx}  
\usepackage{amsfonts}
\usepackage{amsmath}
\usepackage{algorithm}
\usepackage{algorithmic}
\usepackage{multirow}
\usepackage{graphicx}
\usepackage{booktabs}       
\usepackage{nicefrac}       
\usepackage{microtype}      
\usepackage{graphicx,float,url,multirow,makecell,booktabs,color,subfigure,caption,setspace}
\usepackage{times}
\usepackage{graphicx}
\usepackage{amsmath}
\usepackage{amssymb}
\usepackage{subfig}
\newcommand{\namecite}[1]{\citeauthor{#1}~(\citeyear{#1})}

\begin{document}
%
\title{Learning Approximate Stochastic Transition Models}
\author{\textbf{Yuhang Song}$^1$, Christopher Grimm$^2$, Xianming Wang$^3$, \textbf{Michael L. Littman}$^4$ \\
  $^{1}$\textbf{Beijing University of Aeronautics and Astronautics}, $^{2}$\textbf{University of Michigan}, \\ $^{3}$\textbf{Renmin University of China}, $^{4}$\textbf{Brown University}\\
  yuhangsong@buaa.edu.cn, crgrimm@umich.edu, wxm@ruc.edu.cn, michael\_littman@brown.edu}
\maketitle
\begin{abstract}
We examine the problem of learning mappings from state to state, suitable for use in a model-based reinforcement-learning setting, that simultaneously generalize to novel states and can capture stochastic transitions. We show that currently popular generative adversarial networks struggle to learn these stochastic transition models but a modification to their loss functions results in a powerful learning algorithm for this class of problems.
\end{abstract}

\section{Introduction and Background}

Model-based approaches separate the reinforcement-learning (RL) problem into two components. The first component learns a transition model that predicts the next state from the current state and action. The second component uses that model to make decisions by looking ahead to predict the consequences of different courses of actions. This paper focuses on the first problem of acquiring the model, specifically addressing the development of a mechanism for learning to approximate a stochastic transition function.


A Markov decision process (MDP) model of an environment consists of a set of states $S$ and actions $A$, a transition function $T:S\times A \rightarrow \Pi(S)$ mapping state--action pairs to a probability distribution over next states, and a reward function $R:S\times A \rightarrow \Re$.

Since the focus of this paper is not on decision making but on learning the dynamics, we simplify the transition function to $T(\bar{x},a,x')=\mathbb{P}^{\bar{x}}_{r}(x')$, which represents the probability that state $x'$ will follow $\bar{x}$. A separate function can be learned for each action $a\in A$. Although some authors have found there to be an advantage to representing the transitions jointly for all actions~\cite{oh15}, this issue is orthogonal to the representation issue we address here.

To review methods for learning $\mathbb{P}^{\bar{x}}_{r}(x')$, we begin by separating out three representations for transition models. A \emph{query}
model is one that can answer, for any $\bar{x},x'$ pair, the probability of $x'$ given $\bar{x}$. Such a model can be represented as a table if the state space is relatively small~\cite{kearns02}. It can also be captured by a dynamic Bayesian network~\cite{kearns99d,degris2006learning}.
Some types of planners, such as ones based on policy iteration~\cite{Puterman94}, require access to these probabilities to compute expected values.
Query models can be very challenging to work with and learn when the size of the state space if enormous, however, because looping over all the possible values of $x'$ can be too expensive. It is especially problematic when most $\bar{x},x'$ pairs have zero probability, since considering each of them is expensive and pointless.

A \emph{sparse} model is a refinement of the query model that takes a state $\bar{x}$ as input and returns a list of states $N(\bar{x})$ such that $x' \in N(\bar{x})$ if and only if $\mathbb{P}^{\bar{x}}_{r}(x')>0$. Such a representation can be used much more efficiently as it only needs to consider the non-zero entries of $\mathbb{P}^{\bar{x}}_{r}(x')$. Tabular and DBN methods can be used in this setting, but a general approach has yet to be articulated. In addition, they provide no advantage over the query model in environments in which $|N(\bar{x})|$ is intractably large.

An alternative is a \emph{generative} model, which is a function $G$ that, given, $\bar{x}$, produces $x'$.
%
%
Learning a generative model is closely related to classical supervised learning problems. Given examples of $\bar{x},x'$ pairs, they learn a mapping such that $G(\bar{x})$ produces $x'$. When the target mapping is deterministic, many learning algorithms can be brought to bear to learn the transition model~\cite{atkeson97b}.
These learning algorithms can be applied to stochastic transition models, but, as we show, there are significant pitfalls to doing so. An exception is in control problems where the transition model has the form $x' \sim G(x) + \eta$, where the $\eta$ is state independent and typically small and zero-mean noise, so that planning using $G(x)$ results in an approximation to planning with the stochastic model---the noise can be safely ignored during planning~\cite{Bradtke93}.


In our work\footnote{Code to reproduce this work is publicly available online for facilitating future research: https://github.com/YuhangSong/SGAN.}, we capture the transition function adopting a generative adversarial network (GAN) perspective~\cite{goodfellow2014generative}. In the next section, we present GANs and derive our novel variant that is more effective at capturing detailed probability distributions. We provide empirical comparisons between GANs and other approaches to learning, showing that our GAN approach can learn to generalize probabilistic functions effectively.


\section{Modeling Stochastic Transitions with GANs}

The GAN approach to modeling stochastic transition functions focuses on creating a \textit{generator} $G$, which takes in current state $\bar{x}$ and random noise $\mathfrak{n}$ and generates a possible next state $x_g$,
\begin{equation}\label{g-model}
    x_g = G_{\mu}(\bar{x},\mathfrak{n}),
\end{equation}
where $\mu$ is a set of parameters defining $G$ to be set by learning. We assessed the error in $G$ by the L1 distance between the distribution produced by $G$ and the true distribution\footnote{We choose L1 because it is common to measure the error in transition functions this way---a bound in L1 error can be translated to a bound in the reward obtained via the simulation lemma~\cite{kearns02}. Of course, other measures of distribution difference are also valid.}.
A second model $D$, the \textit{discriminator}, takes in current state $\bar{x}$ and next state $x'$. Here, $x'$ is either a state $x_g$ generated by $G$ or a real state $x_r$ from the observed data. The output of the discriminator is interpreted as a score of whether it believes $x' = x_r$. We write
\begin{equation}\label{d-model}
    D_{\theta}(\bar{x},x'),
\end{equation}
where $\theta$ is the parameters defining $D$.

\subsection{WGANs and GP-WGANs}

In the Wasserstein GAN or WGAN~\cite{arjovsky2017wasserstein}, the generator and discriminator attempt to minimize a metric known as the Earth Mover's distance between the generated probability distribution and the real probability distribution:
\begin{equation}\label{eq:em-dst}
W(\mathbb{P}_r, \mathbb{P}_g) = \sup_{\|f\|_L \leq 1} \mathbb{E}_{x_r\sim \mathbb{P}_r}\left[ f(x_r) \right] - \mathbb{E}_{x_g\sim \mathbb{P}_g}\left[ f(x_g) \right],
\end{equation}
where $\| \cdot \|_L \leq 1$ denotes the space of $1$-Lipschitz functions.

Computing the supremum over $1$-Lipschitz functions in Equation~\eqref{eq:em-dst} is computationally intractable. However, \namecite{arjovsky2017wasserstein} showed how to approximate this operation by optimizing the function $f$ as the discriminator\footnote{Because the function assigns scores, it is sometimes referred to as the critic instead of the discriminator.} $D^W_\theta$ and maximizing the expression over parameters $\theta$. This choice leads to the following expression, which encapsulates the optimization procedure for the discriminator and generator networks in the conditional setting:
\begin{equation}
\min_{\mu} \max_{\theta : \|D^W_\theta\|_L \leq 1} \mathbb{E}_{x_r \sim \mathbb{P}_r} \left[ D^W_{\theta}(\bar{x}, x_r)\right] - \mathbb{E}_{x_g \sim \mathbb{P}_{G_\mu}} \left[ D^W_{\theta}(\bar{x}, x_g) \right].
\end{equation}
In words, we are looking for the generator such that even the most discriminating discriminator is unable to assign real data high scores and generated data low scores---the generated and real data are indistinguishable. The training procedure for this loss is performed by taking gradient steps to optimize the discriminator while holding the generator's parameters fixed, and the generator while holding the discriminator's parameters fixed.

Notice that the optimization of $\theta$ requires that $D^W_\theta$ is $1$-Lipschitz throughout the process. The most popular way of enforcing this constraint is by introducing a penalty term~\cite{gulrajani17} in the discriminator's optimization steps, giving the updated expression:
\begin{equation}
\begin{aligned}
&\min_{\mu} \max_{\theta : \|D^W_\theta\|_L \leq 1} \mathbb{E}_{x_r \sim \mathbb{P}_r} \left[ D^W_{\theta}(\bar{x}, x_r)\right] - \mathbb{E}_{x_g \sim \mathbb{P}_{G_\mu}} \left[ D^W_{\theta}(\bar{x}, x_g) \right] \\
&- \lambda \mathbb{E}_{x_\tau \sim \mathbb{P}_{x_\tau}} \left[ (\|\nabla_{x_\tau} D^W_\theta(\bar{x}, x_\tau) \| - 1)^2 \right],
\end{aligned}
\end{equation}
where $\lambda > 0$ is a hyperparameter controlling how strongly to enforce the penalty term and $x_\tau$ is a point drawn from somewhere in the space of $\mathbb{P}_r$ or $\mathbb{P}_{G_\mu}$. Specifically, $x_\tau$ is generated as an interpolation between a pair of real and generated samples: $x_\tau = \tau x_r + (1 - \tau) x_g$ with $x_r \sim \mathbb{P}_r$, $x_g \sim \mathbb{P}_{G_\mu}$ and $\tau \sim [\text{U}](0,1)$. For a detailed derivation of this loss term, see Lemma~1 of \namecite{gulrajani17}. This method, because it combines a gradient penalty with the WGAN, is known as GP-WGAN.

\begin{algorithm}
    \caption{SGAN Training Algorithm for learning to match an observed probability distribution. Default values: number of discriminator iterations per generator iteration $C=5$; batch size $B=32$; hyper-parameter $\delta=0.3$; $l=128$}
    \label{algorithm-table-sgan}

    \textbf{Require:} $C$, $B$, $\alpha$, $\beta_1$, $\beta_2$, $\delta$, dataset containing multiple transition pairs $(\bar{x},x_r)$.

    \textbf{Require:} Initial parameters $\theta_0$ for discriminator $D_{\theta}^S$, initial parameters $\mu_0$ for generator $G_\mu$.
\begin{algorithmic}[1]
    \WHILE{$\theta$ has not converged}
        \FOR{$c=1,\cdots,C$}
            \FOR{$b=1,\cdots,B$}
                \STATE Sample real transition pair $(\bar{x},x_r)$ from dataset
                \STATE Sample noise vector $\mathfrak{n} \sim [\text{U}][0,1]^l$
                \STATE $x_g\leftarrow G_{\mu}(\bar{x},\mathfrak{n})$ //generate next state
                \STATE $T_b \leftarrow \frac{\|x_r-x_g\|}{\delta}$
                \FOR{$t=1,\cdots,T_b$}
                    \STATE Sample $\tau \sim [\text{U}][0,1]$
                    \STATE $x_\tau \leftarrow \tau x_{r} + (1-\tau) x_g$
                    \STATE $L_D^{(b,t)} \leftarrow (\|\nabla_{x_{\tau}}D_{\theta}^S(\bar{x},x_{\tau})-\frac{x_{r}-x_{g}}{\|x_{r}-x_{g}\|}\|)^2$
                \ENDFOR
            \ENDFOR
            \STATE $\theta \leftarrow \theta -  \nabla_{\theta}\frac{1}{B}\sum_{b=1}^{B}\frac{1}{T_b}\sum_{t=1}^{T_b}L_D^{(b,t)}$
        \ENDFOR
        \FOR{$b=1,\cdots,B$}
            \STATE Sample noise vector $\mathfrak{n} \sim [\text{U}][0,1]^l$.
            \STATE $L_G^{b} \leftarrow -D_{\theta}^S(G_{\mu}(\bar{x},\mathfrak{n}))$
        \ENDFOR
        \STATE $\mu \leftarrow \mu- \nabla_{\mu}\frac{1}{B}\sum_{b=1}^{B}L_G^{b}$
    \ENDWHILE
\end{algorithmic}
\end{algorithm}

\subsection{SGANs}

As we demonstrate, WGAN and GP-WGANs struggle to learn generators that closely match the observed transition data. In response, we propose our SGAN in Algorithm~\ref{algorithm-table-sgan}. Its basic steps are:
\begin{itemize}
    \item Sample a real transition pair $(\bar{x},x_r)$ from the dataset.
    \item Generate a transition pair $(\bar{x},x_g)$ with $x_g=G_{\mu}(\bar{x},\mathfrak{n})$, where the components of the noise vector $\mathfrak{n}$ are drawn from $[\text{U}][0,1]$ and $G_{\mu}$ denotes the generator with parameter $\mu$.
    \item Train the discriminator $D^S_\theta$ with parameter $\theta$ to minimize $L^{\textrm{SGAN}}_D$, which we will define later.
    \item Train the generator $G_\mu$ with its loss $L_G=\mathbb{E}_{\mathfrak{n}\sim [\text{U}][0,1]}\{-D_{\theta}^S (G_{\mu}(\bar{x},\mathfrak{n}))\}$.
\end{itemize}

Apart from the new loss function $L^{\textrm{SGAN}}_D$, the above procedure is shared with that of WGANs and GP-WGANs.

Now, we define the new discriminator loss function $L^{\textrm{SGAN}}_D$,
\begin{eqnarray}\label{loss-d-in-sgan}
    L^{\textrm{SGAN}}_D &=& \mathbb{E}_{x_r\sim\mathbb{P}_r,x_g\sim\mathbb{P}_g,\tau\sim [\text{U}][0,1]} \nonumber\\
    && \; \left[ (\|\nabla_{x_{\tau}}D_{\theta}^S(\bar{x},x_{\tau})-\frac{x_{r}-x_{g}}{\|x_{r}-x_{g}\|}\|)^2 \right],
\end{eqnarray}
where $x_\tau$ is
\begin{equation}\label{x-tau}
 x_{\tau} = \tau x_{r} + (1-\tau) x_g,
\end{equation}
\begin{equation}\label{tau}
 \tau \sim [\text{U}][0,1].
\end{equation}
But, unlike the way $x_\tau$ is sampled in the GP-WGAN, the SGAN samples $x_\tau$ for $T$ times given each $x_r, x_g$ pair. The value $T$ is computed by
\begin{equation}\label{T-method}
    T=\frac{\|x_r-x_g\|}{\delta}.
\end{equation}
Here, $\delta$ is a hyper-parameter of the algorithm, the choice of which is discussed in the experiment section.

\section{Training the discriminator in SGANs}

By executing Algorithm~\ref{algorithm-table-sgan}, our $D_{\theta}^S$ is modelling a different discriminator from the $D_{\theta}^W$ in WGAN---we refer to our new discriminator as the SGAN discriminator. Following the derivation of the optimal WGAN discriminator, we now express the optimal SGAN discriminator, denoted by $D^{S^\ast}$:
\begin{eqnarray}\label{opt-d-sgan}
    D^{S^\ast}(x) = \int^x_{0} ( \int_{x}^{1} \mathbb{P}_r(\hat{x}) d \hat{x} - \int_{x}^{1} \mathbb{P}_g(\hat{x}) d \hat{x} ) d x. \nonumber
\end{eqnarray}

This section focuses on proving the SGAN algorithm can minimize $D^{S^\ast}$. Our argument proceeds in two steps:
\begin{itemize}
    \item Lemma~\ref{lemma} shows an important property arising from sampling $x_\tau$ for $T$ times in the SGAN algorithm.
    \item Building on this property, Theorem~\ref{theorem} shows that the loss function of $D$ in the SGAN algorithm (Equation~\eqref{loss-d-in-sgan}) leads to $D^{S^\ast}$.
\end{itemize}
We restrict our argument to a one-dimensional setting for simplicity even though the algorithm is implemented in tested in high dimensional problems.

\newtheorem{theorem}{Theorem}
\newtheorem{lemma}{Lemma}

\begin{lemma}
Consider an event denoted by: $x_\tau\overset{T}{=}x_n$, defined to means we sample $x_\tau$ $T$ times and the $x_\tau = x_n$ at least once. To be clear, $x_\tau$ $x_r$, $x_g$ are all random variables while $x_n$ represents a specific fixed value. Assuming
\begin{equation}\label{T}
  T = |x_r-x_g|/{\delta},
\end{equation}
it follows that
\begin{eqnarray}
\lefteqn{P(x_\tau\overset{T}{=}x_n|x_r,x_g)} \\
    &=&
    \begin{cases}
        c &\mbox{$x_r<x_n<x_g,x_g<x_n<x_r$}\\
        0 &\mbox{else},
    \end{cases}
\end{eqnarray}
where $c$ is a constant independent of the values of the random variables.
\label{lemma}
\end{lemma}

\noindent \textbf{Proof of Lemma~\ref{lemma}:} We begin with a derivation in which we have discretized the one-dimensional space into intervals of size $\varepsilon$. Notationally, the discretized versions of the variables are marked with a check over the variable name. Later on, we will derive what happens to these expressions as we take the limit of $\varepsilon\rightarrow 0$, bringing us back to statements about continuous space. We have
\begin{equation}\label{p-epsilon-pr-pg}
    P(\check{x}_\tau\overset{1}{=}\check{x}_n|\check{x}_r,\check{x}_g)=
    \begin{cases}
        \frac{1}{|\check{x}_r-\check{x}_g|/\varepsilon} &\mbox{$\check{x}_r<\check{x}_n<\check{x}_g,\check{x}_g<\check{x}_n<\check{x}_r$}\\
        0 &\mbox{otherwise}.
    \end{cases}
\end{equation}
If we sample $\check{x}_\tau$ for $T$ times,
\begin{eqnarray}\label{p-tau}
\lefteqn{ P(\check{x}_\tau\overset{T}{=}\check{x}_n|\check{x}_r,\check{x}_g)}\\
    &=& 1 - (1-P(\check{x}_\tau\overset{1}{=}\check{x}_n|\check{x}_r,\check{x}_g))^T \nonumber\\
    &=&
    \begin{cases}
        1 - (1-\frac{1}{d/\varepsilon})^{d/\delta} &\mbox{$\check{x}_r<\check{x}_n<\check{x}_g,\check{x}_g<\check{x}_n<\check{x}_r$} \\
        0 &\mbox{otherwise}. \nonumber
    \end{cases}
\end{eqnarray}
We relate $\varepsilon$ to $\delta$ via a positive integer multiple\footnote{This is true when consider $\varepsilon \rightarrow 0$ is the minimal value a computer can operate.} $z$:
\begin{equation}\label{deltal-satisfy}
  \delta = z \varepsilon,
\end{equation}
where $z \in Z^{+}$.
To connect back to Equation~\eqref{p-tau}, we consider following limit,
\begin{eqnarray}\label{limit-constant}
\lefteqn{\lim_{\delta=z\varepsilon,\varepsilon\rightarrow 0} (1-\frac{1}{d/\varepsilon})^{d/\delta}} \nonumber\\
    \nonumber &=& \lim_{\delta=z\varepsilon,\varepsilon\rightarrow 0} e^{d/\delta \ln(1-\frac{1}{d/\varepsilon})} \\
    \nonumber &=& \lim_{\delta=z\varepsilon,\varepsilon\rightarrow 0} e^{\frac{\ln(\frac{d-\varepsilon}{d})}{\delta/d}} \\
    \nonumber &=& \lim_{\varepsilon\rightarrow0} e^{\frac{\frac{d}{d-\varepsilon}\frac{-1}{d}}{z/d}} \\
    &=& e^{-1/z} \label{e:e}
\end{eqnarray}
Substituting Equation~\eqref{limit-constant} into Equation~\eqref{p-tau} and taking the limit as $\varepsilon\rightarrow 0$, we have
\begin{eqnarray}\label{final-p-inter}
\lefteqn{P(x_\tau\overset{T}{=}x_n|x_r,x_g)}\\
&=& \lim_{\delta=z\varepsilon,\varepsilon\rightarrow 0} P(\check{x}_\tau\overset{T}{=}\check{x}_n|\check{x}_r,\check{x}_g) \nonumber\\
    &=&
    \begin{cases}
        1 - e^{-1/z} &\mbox{$x_r<x_n<x_g,x_g<x_n<x_r$}\\
        0 &\mbox{otherwise},
    \end{cases}
\end{eqnarray}
where we have switched back to the continuous space and finished the proof.

\begin{theorem}\label{theorem}
Under all the assumptions in Lemma~\ref{lemma}, if we update $D^S_\theta$ with loss
\begin{equation}\label{loss-d-in-sgan-1}
    L = (|\nabla_{x_{\tau}}D_{\theta}^S(\bar{x},x_{\tau})-\frac{x_{r}-x_{g}}{|x_{r}-x_{g}|}|)^2,
\end{equation}
then $D_{\theta}^S(x)$ approaches
\begin{eqnarray}\label{explain-d}
    D_{\theta}^S (x) = c \, D^{S^\ast}(x),
\end{eqnarray}
for an undefined constant $c$.
\end{theorem}

\noindent \textbf{Proof of Theorem~\ref{theorem}:}
Equation~\eqref{loss-d-in-sgan-1} encourages $\nabla_{x_{\tau}}D_{\theta}^S(x_{\tau})$ to approach $\frac{x_{r}-x_{g}}{|x_{r}-x_{g}|}$. Since $x_{r}$ and $x_{g}$ are random variables, $\nabla_{x_{\tau}}D_{\theta}^S(x_{\tau})$ is updated toward $+1$ and $-1$ randomly. As a result, we should consider the learned value of $\nabla_{x_{\tau}}D_{\theta}^S(x_{\tau})$ as it relates to the probability that it gets different updates.
Let us take a look at $\nabla_{x_{\tau}}D_{\theta}^S(x_{\tau})$ at an arbitrary point $x_n$:
\begin{eqnarray}\label{d-at-xn}
\lefteqn{\mathbb{E}_{x_r\sim\mathbb{P}_r,x_g\sim\mathbb{P}_g, \tau \sim [\text{U}][0,1]} \left[ \nabla_{x_{\tau}=x_n} D_{\theta}^S(x_{\tau}=x_n,\delta) \right]} \nonumber\\
    &=& \mathbb{E}_{x_r\sim\mathbb{P}_r,x_g\sim\mathbb{P}_g, \tau \sim [\text{U}][0,1]} \left[\frac{x_{r}-x_{g}}{|x_{r}-x_{g}|}\right] \nonumber\\
    &=& P(x_\tau\overset{T}{=}x_n|x_g<x_n<x_r) P(x_g<x_n<x_r) \\
    && - P(x_\tau\overset{T}{=}x_n|x_r<x_n<x_g) P(x_r<x_n<x_g).\nonumber
\end{eqnarray}
From Lemma~\ref{lemma}, we know that
\begin{equation}\label{p-inter-conditional-1}
  P(x_\tau\overset{T}{=}x_n|x_g<x_n<x_r)=c
\end{equation}
and
\begin{equation}\label{p-inter-conditional-2}
  P(x_\tau\overset{T}{=}x_n|x_r<x_n<x_g)=c,
\end{equation}
for some hyper-parameter controlled constant $c$. In the context of Equation~\eqref{d-at-xn}, we have
\begin{eqnarray}\label{explain-d-gradient}
\lefteqn{\nabla_{x_{\tau}=x_n} D_{\theta}^S(x_{\tau}=x_n)} \nonumber\\
    &=& [P(x_g<x_n<x_r) - P(x_r<x_n<x_g)]c \nonumber\\
    &=& [P(x_g<x_n)P(x_n<x_r) \nonumber\\
    && -  P(x_r<x_n)P(x_n<x_g) ]c \nonumber\\
    &=& [ \int_{0}^{x_n}\mathbb{P}_g(x)dx \int_{x_n}^{1}\mathbb{P}_r(x)dx  \nonumber\\
    && - \int_{0}^{x_n}\mathbb{P}_r(x)dx \int_{x_n}^{1}\mathbb{P}_g(x)dx ]c\nonumber\\
    &=& \left[\int_{x_n}^{1}\mathbb{P}_r(x)dx-\int_{x_n}^{1}\mathbb{P}_g(x)dx\right]c.
\end{eqnarray}
Further,
based on Equation~\eqref{explain-d-gradient},
\begin{eqnarray}
\lefteqn{\mathbb{E}_{x_r\sim\mathbb{P}_r,x_g\sim\mathbb{P}_g, \tau \sim [\text{U}][0,1]} \left[ D_{\theta}^S(x_{\tau}=x_n) \right]} \nonumber\\
    &=& \left[\int^{x_n}_{0} \left( \int_{x}^{1} \mathbb{P}_r(\hat{x}) d \hat{x} - \int_{x}^{1} \mathbb{P}_g(\hat{x}) d \hat{x} \right) d x \right] c \nonumber\\
    &=& \left[ D^{S^\ast}(x_{\tau}=x_n) \right] c,
\end{eqnarray}
completing the proof.

\section{Experiments}

This section presents experimental results.

\subsection{Comparison Algorithms}

We compare SGAN against a tabular learner, a deterministic deep network and the state-of-the-art\footnote{In preliminary work, we evaluated GAN and WGAN and found they were consistently worse than GP-WGAN.} GP-WGAN.

Given a set $\mathbb{S}$ of $\langle \bar{x},x_r\rangle$ samples, our tabular learner simply memorizes all of them. It then estimates $\mathbb{P}^{\bar{x}}_{r}(x') = |\{\langle \bar{x},x'\rangle \in \mathbb{S}\}|/ |\{ \langle \bar{x},\cdot\rangle \in \mathbb{S}\}$. If $\bar{x}$ was not observed, it returns a default value that is interpreted as an error in our experiments.

For all deep neural network based methods, that is, deterministic deep network, GP-WGAN and SGAN, we used the same Adam optimizer with parameters $\alpha=0.0001$, $\beta_1=0.0$, and $\beta_2=0.9$. We kept the network structure as uniform as possible.  For all 3D convolutional neural networks (C) layers and 3D deconvolutional neural networks (DC) layers~\cite{ji20133d}, we used LeakyReLU activation with a negative slope of $0.001$ and kernel size $D\times4\times4$, stride $1\times2\times2$, and padding $0\times1\times1$ (sizes are sizes reported as $\text{Depth}\times\text{Height}\times\text{Width}$).
We denote a C layer as $[\text{C}^D]$, where $D$ denotes different kernel depth $D\times4\times4$. Similarly, we denote a DC layer as $[\text{DC}^D]$. For all fully connected (F) layers, we used LeakyReLU activation with a negative slope of $0.001$. We denote a F layer mapping size $a$ to size $b$ as $\text{F}^{a\rightarrow b}$.

To be able to precisely describe the networks in our experiments,  we define a few special terms:
\begin{itemize}
    \item \textit{Squeeze layer} [\text{S}]. This layer always appears after a C layer, and it first flattens the output of the C layer, then uses a F layer mapping the flattened vector to 512.
    \item \textit{Concatenate Layer} [\text{CL}]. This layer always appears after a F layer of size 512. It concatenates the output of the F layer with noise vector $\mathfrak{n}$, which means it is a F layer mapping from (512+$l$) to 512, and the dimensionality of the noise vector is $l=128$. For uniformity, we run deterministic deep networks using the same structure, replacing $\mathfrak{n}$ with a zero vector of the same size.
    \item \textit{Unsqueeze layer} [\text{U}]. This layer appears after a F layer of size 512 and before a DC layer. It first uses a F layer that maps 512 to the size of the input of the following DC layer, then reshapes the output vector to the shape of the input of the following DC layer.
    \item \textit{Linear output layer} [\text{L}]. This layer appears after a F layer of size 512. It is a linear layer that maps 512 to 1.
    \item \textit{Layer sequence}. We use arrows to show how layers are connected: $[\ast]\rightarrow[\ast]\rightarrow[\ast]$.
\end{itemize}

\subsubsection{Network structure for all grid domains.}
For vector-based domains, we denote the size of the vector as $V$. Our deterministic deep network, $G$ of GP-WGAN and $G$ of SGAN used the same structure:
\begin{itemize}
    \item $[\text{F}^{V\rightarrow 512}]\rightarrow[\text{F}^{512\rightarrow 512}]\rightarrow[\text{S}]\rightarrow[\text{CL}]\rightarrow[\text{U}]\rightarrow[\text{F}^{512\rightarrow 512}]\rightarrow[\text{F}^{512\rightarrow V}]$
\end{itemize}
For vector-based domains, $D$ of GP-WGAN and $D$ of SGAN use the same structure of:
\begin{itemize}
    \item $[\text{F}^{2V\rightarrow 512}]\rightarrow[\text{F}^{512\rightarrow 512}]\rightarrow[\text{S}]\rightarrow[\text{L}]$
\end{itemize}
For image representations, deterministic deep network, $G$ of GP-WGAN and $G$ of SGAN, used the same structure of:
\begin{itemize}
    \item $[\text{C}^1]\rightarrow[\text{C}^1]\rightarrow [\text{S}]\rightarrow[\text{CL}]\rightarrow [\text{U}]\rightarrow[\text{DC}^1]\rightarrow[\text{DC}^1]$
\end{itemize}
For image representations, $D$ of GP-WGAN and $D$ of SGAN used the same structure of:
\begin{itemize}
    \item $[\text{C}^2]\rightarrow[\text{C}^1]\rightarrow[\text{S}]\rightarrow[\text{L}]$
\end{itemize}

\subsubsection{Network structure for marble domain.}
The deterministic deep network, $G$ of GP-WGAN, and $G$ of SGAN used the same structure of:
\begin{itemize}
    \item $[\text{C}^2]\rightarrow[\text{C}^1]\rightarrow[\text{C}^1]\rightarrow[\text{S}]\rightarrow[\text{CL}]\rightarrow[\text{U}]\rightarrow[\text{DC}^1]\rightarrow[\text{DC}^1]\rightarrow[\text{DC}^1]$
\end{itemize}
Networks $D$ of GP-WGAN and $D$ of SGAN used the same structure of:
\begin{itemize}
    \item $[\text{C}^2]\rightarrow[\text{C}^2]\rightarrow[\text{C}^1]\rightarrow[\text{S}]\rightarrow[\text{L}]$
\end{itemize}

\subsection{SGAN Parameter Selection}

The SGAN hyper-parameter $\delta$ sets a weak trade-off between training speed and optimizing the SGAN objective function. Smaller values of $\delta$ will guarantee SGAN is achieved but result in slow training speed as many $x_\tau$ values end up being sampled. We found that, as long as $T$ is kept above $3$, $\delta$ can be made as small as desired. As such, we recommend keeping $T$ around $3$, since it reduces computational cost without sacrificing suboptimality. In our experiments, we used $\delta=0.3$ for all vector-based domains and $\delta=1$ for all image-based domains.

When setting $\delta$ in a new domains, we recommend a simple way to find the maximal $\delta$ that keeps $T$ above $3$. If the state can be decomposed into a deterministic piece (like maintaining the background) and a stochastic piece (like deciding which local move to make), let $s^d$ be a state that highlights the deterministic aspect (such as a pure black background in our 2D grid domain) and let $s^s$ be a state that highlights the stochastic aspect (such as a black background with a white agent on it), and set $\delta=\frac{\|s^d-s^s\|}{3}$.

\subsection{Evaluation metric}

As has been mentioned, we evaluate the error in $G$ by the L1 distance between the distribution produced by $G$ and the true distribution.  In evaluating the distributions, we found that the GANs sometimes produce states that are not meaningful in the context of the domain under study. In these cases, we also include an evaluation of \emph{sample validity}---how often the model produces meaningful samples. When possible, we resample after an invalid sample is produced. Our rule for separating valid from invalid samples is to call an output valid if its pixel-wise deviation to a corresponding real state is less than 0.1.

To make comparisons as fair as possible, all methods are evaluated after being trained with 100,000 iterations with a batch size of 32 for each iteration. For all grid domains represented by images, we used a block size of 4, which means every cell in the grid is $4\times4$ pixels.

\subsection{Simple Domains}

\begin{table*}\renewcommand{\arraystretch}{1.1}
\centering
\resizebox{\textwidth}{!}{%
\begin{tabular}{cccccccc}
\hline
L1 Loss/Sample Validity & Representation & Size & Dynamic & Tabular learner & Deterministic deep network & GP-WGAN & SGAN \\ \hline
\multirow{6}{*}{1D Grid} & \multirow{3}{*}{Vector} & 5 & 1/3:2/3 & \textbf{0.001/100\%} & /0\% & 0.231/92\% & 0.046/99\% \\
 &  & 10 & 1/3:2/3 & \textbf{0.001/100\%} & /0\% & 0.103/99\% & 0.038/99\% \\
 &  & 20 & 1/3:2/3 & \textbf{0.001/100\%} & /0\% & 0.089/98\% & 0.035/98\% \\ \cline{2-8}
 & \multirow{3}{*}{Image} & 5 & 1/3:2/3 & \textbf{0.001/100\%} & /0\% & 0.149/97\% & 0.054/97\% \\
 &  & 10 & 1/3:2/3 & \textbf{0.001/100\%} & /0\% & 0.221/94\% & 0.106/97\% \\
 &  & 20 & 1/3:2/3 & \textbf{0.001/100\%} & /0\% & 0.152/94\% & 0.076/93\% \\ \hline
\multirow{2}{*}{2D Grid} & \multirow{2}{*}{Image} & 5 & 0.8:0.1:0.0:0.1 & \textbf{0.018/100\%} & /0\% & 0.180/92\% & 0.109/92\% \\
 &  & 5 & 0.25:0.25:0.25:0.25 & \textbf{0.018/100\%} & /0\% & 0.450/77\% & 0.082/90\% \\ \hline
\multicolumn{4}{c}{Overall} & \textbf{0.005/100\%} & /0\% & \multicolumn{1}{l}{0.196/93\%} & \multicolumn{1}{l}{0.068/96\%} \\ \hline
\end{tabular}%
}
\caption{Results on Simple Domains.}
\label{table-simple}
\end{table*}
\vspace{-0.2em}

We first investigate SGAN on two simple domains. The state in the \emph{1D Grid} domain is the location of an agent in a $1\times n$ hallway. The dynamics are that the agent transitions with a $\frac{1}{3}$ probability to the left and a $\frac{2}{3}$ probability to the right right. If a transition would cause the agent to exit the $1\times n$ hallways, it remains in place. We represented the state of the domain in two different ways. In the vector representation, the learner was presented an $n$-bit vector with a 1 at the location of the agent. The image representation was similar, except an entire $4\times 4$ collection of pixels replaced each bit position in the vector representation. We ran experiments for all $n \in \{5, 10, 20\}$.

States in the \emph{2D Grid} domain consisted of two-dimensional images with a grid size of $5\times 5$ (and therefore an input size of $20\times 20$). We examined two different transition dynamics for these grids: a uniform random walk on the 4 cardinal directions ($0.25:0.25:0.25:0.25$) and the Russell-Norvig~\cite{russell94} grid dynamics ($0.8:0.1:0.0:0.1$) with the intended movement direction being north. Any transition that would take the agent out of the grid resulted in no state change.

As shown in Table~\ref{table-simple}, our SGAN performs well for most of the simple domains. The average improvement of SGAN over GP-WGAN was $0.128$ on L1 loss and $3\%$ on sample validity.

Unsurprisingly, SGAN performed worse than the tabular learners in these domains---the tabular method is effectively matching ground truth in these cases as there is more than enough data to observe all possible inputs and accurately estimate their associated outputs.

\begin{figure}
\centering
\begin{tabular}{cc}
\includegraphics[width=0.4\columnwidth]{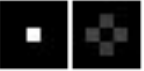}
&
\includegraphics[width=0.4\columnwidth]{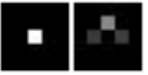}
\\
(a) & (b)
\end{tabular}
\caption{Transition pairs modelled by a deterministic deep network on the $5\times 5$ \emph{2D Grid}, where the image on the left side is the start state, and the image on the right side is the generated next state. (a) The learned transition pair under the uniform random walk dynamics ($0.25:0.25:0.25:0.25$). (b) The learned transition pair under the Russell-Norvig grid dynamics ($0.8:0.1:0.0:0.1$).}
\label{figure-deter}
\end{figure}

\begin{figure}
\centering
\begin{tabular}{ccc}
\includegraphics[width=0.22\columnwidth]{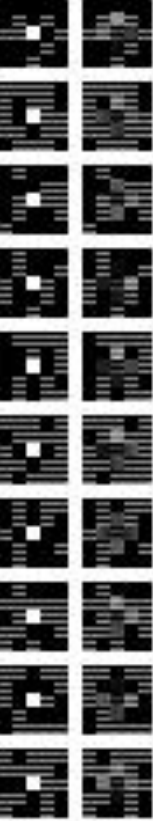}
&
\includegraphics[width=0.22\columnwidth]{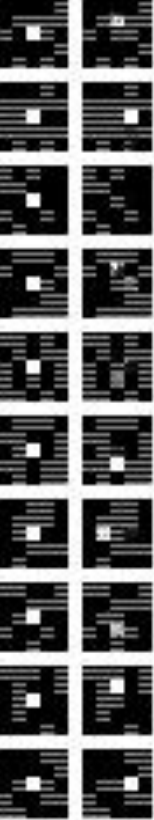}
&
\includegraphics[width=0.22\columnwidth]{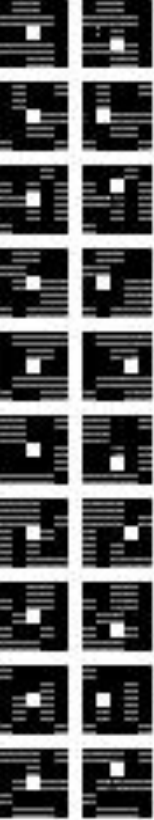}
\\
(a) & (b) & (c)
\end{tabular}
\vspace{-0.8em}
\caption{Transition pairs generated by different models on the $5\times 5$ 2D Grid with Random Backgrounds domain with uniform random walk dynamics ($0.25:0.25:0.25:0.25$). (a) Transition pairs generated by the deterministic deep network. (b) Transition pairs generated by GP-WGAN. (c) Transition pairs generated by SGAN. For every subfigure, the left row of images is the start state with agent fixed to the same position for evaluation but backgrounds chosen at random, and the right row of images is the generated next state based on that start state.}
\label{figure-rbg}
\end{figure}

The deterministic deep net cannot generate any valid samples despite the simplicity of these problems. Figure~\ref{figure-deter} shows what the network learns, which is to ``hedge its bets'' by predicting a fractional transition to each of the neighboring cells. These transitions are sensitive to the transition probabilities, with more likely next positions being given more weight. This kind of output is a known consequence of using a deterministic network with least squared loss to learn a stochastic function.

The GAN networks learn to produce correct next states. We increase the difficulty of these tasks in the next section.

\subsection{Complex Domains}

\begin{table*}\renewcommand{\arraystretch}{1.1}
\centering
\resizebox{\textwidth}{!}{%
\begin{tabular}{cccccccc}
\hline
\begin{tabular}[c]{@{}c@{}}L1 Loss/Sample Validity\end{tabular} & Representation & Size & Dynamic & Tabular learner & Deterministic deep network & GP-WGAN & SGAN \\ \hline
\multirow{2}{*}{\begin{tabular}[c]{@{}c@{}}2D Grid with\\ Obstacles\end{tabular}} & \multirow{2}{*}{Image} & \multirow{2}{*}{5} & 0.8:0.1:0.0:0.1 & \textbf{0.021/100\%} & /0\% & 0.099/96\% & 0.098/97\% \\
 &  &  & 0.25:0.25:0.25:0.25 & \textbf{0.018/100\%} & /0\% & 0.151/92\% & 0.120/94\% \\ \hline
\multirow{2}{*}{\begin{tabular}[c]{@{}c@{}}2D Grid with\\ Random Background\end{tabular}} & \multirow{2}{*}{Image} & 5 & 0.8:0.1:0.0:0.1 & 2.000/100\% & /0\% & 0.255/90\% & \textbf{0.118/93\%} \\
 &  & 5 & 0.25:0.25:0.25:0.25 & 2.000/100\% & /0\% & 0.619/71.4\% & \textbf{0.161/91\%} \\ \hline
\multicolumn{4}{c}{Overall} & 1.009/100\% & /0\% & 0.281/87.35\% & \textbf{0.124/94\%} \\ \hline
Marble & Image & / & / & / & \begin{tabular}[c]{@{}c@{}}Up:Down:Invalid\\ 100\%:0\%:0\%\end{tabular} & \begin{tabular}[c]{@{}c@{}}Up:Down:Invalid\\ 80\%:0\%:20\% \end{tabular} & \textbf{\begin{tabular}[c]{@{}c@{}}Up:Down:Invalid\\ 43\%:34\%:23\%\end{tabular}} \\ \hline
\end{tabular}%
}
\caption{Results on Complex Domains.}
\label{table-complicated}
\vspace{-1.6em}
\end{table*}

We also evaluated our SGAN in three more complicated domains.

The \emph{2D Grid with Obstacle} domain is identical to the 2D Grid domain, except an impassable object is included in the middle of the grid. Note that, in the 2D Grid with Obstacle domain, the agent is represented by setting the corresponding pixel values to be 1 (white), and the obstacle is represented by pixel values of 0.5 (gray).

The \emph{2D Grid with Random Backgrounds} domain is another 2D Grid domain. Unlike the previous domain, obstacles can appear at any location. Each grid cell is represented by two kinds of features---a `fence feature' denoting the presence of an obstacle and an 'agent feature' denoting the presence of the agent.
Since the space of possible backgrounds is enormous, models have to generalize from their limited data set to learn the underlying rules governing the dynamics. The size of the dataset we use for this domain is a $10^{-6}$ fraction of the total number of possible transition pairs. As such, the tabular learner performs extremely poorly in this task.

For these complex domains, we find that it is common for networks to need to learn two things: How to copy the background features and how to capture the probabilistic aspects of the transitions. Learning one of these can interfere with learning the other. A common failure mode is for the output layers of the $G$ network to lose their connection to $\mathfrak{n}$ when it learns the deterministic part of the transition. Once those aspects are learned, it can be difficult to recover the connections to the noise inputs $\mathfrak{n}$.
To encourage the network to retain these connections, we have an optional additional loss in $G$:
\begin{eqnarray}\label{loss-g-n}
    L_G^{\mathfrak{n}}=-\log(1+\|\nabla_{\mathfrak{n}}G_{\mu}(\bar{x},\mathfrak{n})\|),
\end{eqnarray}
where the purpose of the $\log(1+\cdot)$ operation is to restrain from becoming too large. We combine this loss with the original loss of $G$ by simple addition with a weighting coefficient of $\rho=1.0$. We only implement this technique on the three complex domains, as the deterministic part in the transition is relatively complicated and important. In practice, we found this additional loss of $G$ gives a significant improvement on all GAN-related methods.

As shown in Table~\ref{table-complicated}, our SGAN results in improvements in the most complex domains in terms of L1 loss and sample validity.
Once again, the deterministic deep net cannot generate any valid samples. Figure~\ref{figure-rbg} gives a visualization from the deterministic deep network, GP-WGAN, and SGAN on the 2D Grid with Random Backgrounds domain. From this figure, we can see that all the neural network-based methods are able to learn how to copy the background forward, showing their generalization capability. Of the three, however, only GP-WGAN and SGAN can model stochastic transitions, and SGAN does a better job of capturing the mapping.

\subsection{Physical Domain}

In addition to these simulated video-game-like domains, we built a real-world experimental apparatus where stochasticity stems from low-level physical interactions. Our \emph{Marble} domain consists of a self-resetting marble track. We used a video camera to capture the interactions on one particular part of the track consisting of a bowl-like space that is split by a pillar in the middle. We found that the marble, encountering the pillar from the left side, randomly heads in one of two directions, up or down. We collected 259 minutes of video on this single split at 30 frames per second. To focus on the interesting part of the stochastic transitions, we filtered out the frames that did not contain the marble by computing the pixel-wised variance of frames and setting a threshold to judge if there is a marble present. This process left 22,427 transition pairs in total.

\begin{figure}
\centering
\subfigure[Real marble sequence]{
    \includegraphics[width=1.0\columnwidth]{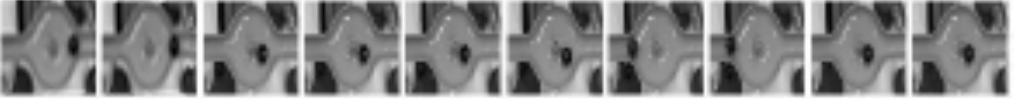}
}
\subfigure[Generated Marble sequence from deterministic deep network.]{
    \includegraphics[width=1.0\columnwidth]{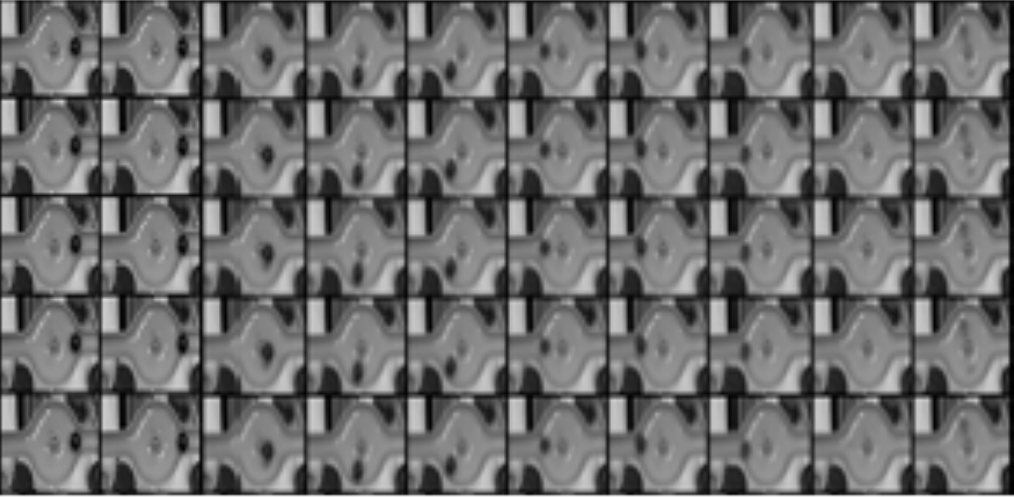}
}
\subfigure[Generated Marble sequence from GP-WGAN.]{
    \includegraphics[width=1.0\columnwidth]{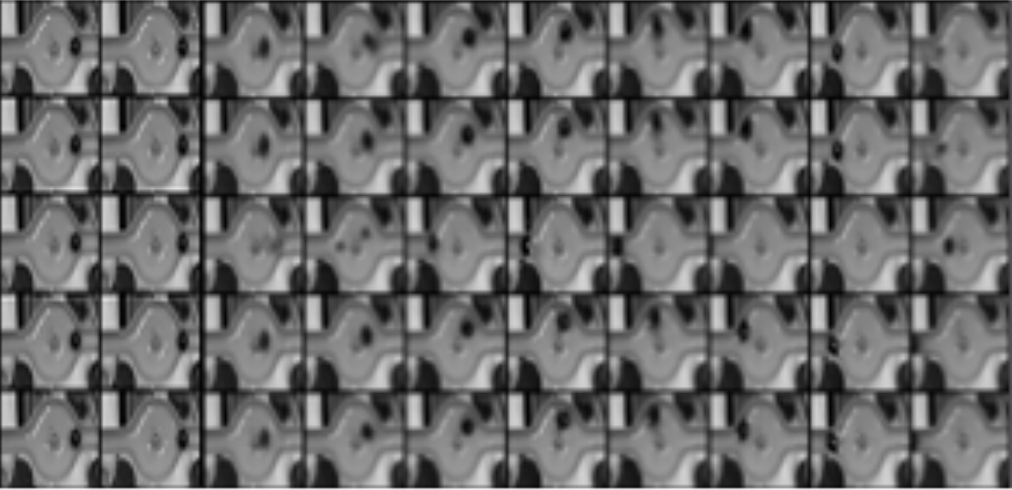}
}
\subfigure[Generated Marble sequence from SGAN.]{
    \includegraphics[width=1.0\columnwidth]{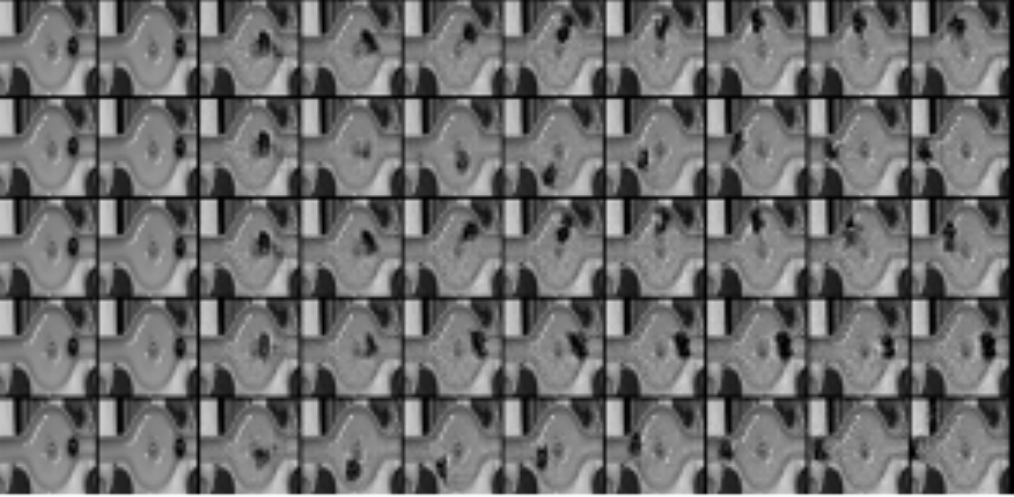}
}
\caption{One real sequence and multiple generated sequences in the Marble domain from the deterministic deep net, GP-WGAN, and SGAN, respectively. Each row is a generated sequence. Time increases from left to right. We report 5 sequences for each method. Each image after the first pair is generated based on previous two images.}
\label{figure-marble}
\end{figure}

To evaluate this domain, the $\bar{x}$ for all methods was set to two consecutive frames and the learner's job was to predict the succeeding frame.
We generate a sequence of marble images starting from two frames. Figure~\ref{figure-marble} shows one real sequence and multiple generated sequences from a deterministic deep net, GP-WGAN, and SGAN, respectively.

As shown in the last row of Table~\ref{table-complicated}, we  manually generated a statistic concerning whether the generated sequences show the marble going up or down or following an invalid path based 30 randomly generated samples.
We can see from these figures that the deterministic deep net can generate valid next states but the sequence is always identical---it fails to model the stochastic transitions. In contrast, GP-WGAN can generate multiple outputs coming from the sequences, but the sequences are still quite deterministic compared to that of SGAN. The SGAN can generate multiple sequences of diverse transitions---it is the best at modeling this stochastic dynamical system.

One caveat is that we note that the generated images from the SGAN are generally of a lower quality than those from the deterministic deep net or GP-WGAN. We currently have no insight into why the SGAN is not generating clear images---we leave it to future research to study in detail.

\section{Conclusion and Future Work}

The SGAN approach provides a promising way of learning transition functions that require both generalization and complex stochastic models. We plan to continue to improve the robustness and accuracy of the method, returning to the original motivation of learning stochastic transitions for model-based reinforcement learning.

\bibliographystyle{aaai}
\bibliography{mlittman}

\end{document}